\documentclass[11pt,a4paper]{article}
\usepackage[utf8]{inputenc}
\usepackage[sc]{mathpazo}
\usepackage[english]{babel}
\usepackage{amsmath}
\usepackage{amsfonts}
\usepackage{amssymb}
\usepackage{graphicx}
\usepackage[left=3cm,right=3cm,top=2cm,bottom=2cm]{geometry}
\usepackage{authblk}
\usepackage[hidelinks]{hyperref}
\usepackage{bm}
\usepackage{adjustbox}
\usepackage[gen]{eurosym}

\usepackage{booktabs}
\newcommand{\fref}[1]{Figure~\ref{#1}}
\newcommand{\tref}[1]{Table~\ref{#1}}
\renewcommand{\eqref}[1]{Equation~\ref{#1}}
\renewcommand{\vec}[1]{\mathbf{#1}}

\title{Determining offshore wind installation times using machine learning and open data}
\date{\today}

\author[1]{Bo Tranberg\thanks{\href{mailto:bo@entolabs.co}{bo@entolabs.co}}}
\author[2]{Kasper Koops Kratmann\thanks{\href{mailto:kasper@i4blades.dk}{kasper@i4blades.dk}}}
\author[3]{Jason Stege\thanks{\href{mailto:jason.stege@siemensgamesa.com}{jason.stege@siemensgamesa.com}}}
\affil[1]{\href{https://ento.ai}{Ento Labs ApS}}
\affil[2]{I4 Blades ApS}
\affil[3]{\href{https://www.siemensgamesa.com}{Siemens Gamesa Renewable Energy A/S}}

\begin{document}
\maketitle

\begin{abstract}\noindent
The installation process of offshore wind turbines requires the use of expensive jackup vessels. These vessels regularly report their position via the Automatic Identification System (AIS). This paper introduces a novel approach of applying machine learning to AIS data from jackup vessels. We apply the new method to 13 offshore wind farms in Danish, German and British waters. For each of the wind farms we identify individual turbine locations, individual installation times, time in transit and time in harbor for the respective vessel. This is done in an automated way exclusively using AIS data with no prior knowledge of turbine locations, thus enabling a detailed description of the entire installation process
\end{abstract}

\tableofcontents

\newpage
\section{Introduction}
The installed capacity of offshore wind has increased by more than a factor of 10 during the last decade \cite{WindEUStats}, and is expected to keep growing in the future \cite{WindEU2020}. Simultaneously, the global weighted average Levelized Cost of Energy (LCOE) for offshore wind decreased by 20\% from 2010 to 2018 \cite{irena2019}. Until recently offshore wind has relied on government subsidies, but with the recent non-subsidized bids driven by cost reductions such schemes are becoming less important \cite{vattenfall2018,orsted2018}. However, potential for cost reduction remains and should be continuously pursued to drive the transition to a fully sustainable energy system.

The installation process of offshore wind requires the use of jackup vessels with day-rates exceeding 100.000 \euro{} \cite{Ahn2017}. These vessels regularly report their position via the Automatic Identification System (AIS), a radio transponder technology developed for real-time vessel tracking to avoid collisions at sea \cite{Mou2010}. The AIS data offers accurate position data at a high temporal resolution and years of data is globally available either publicly or at low cost \cite{ais,marinetraffic}.

A statistical analysis of AIS data has been used to improve collision avoidance at Rotterdam port \cite{Mou2010}. A visualization of ship routes based on historical AIS data has been proposed as a way to determine the level of traffic at locations of interest as well as identify e.g. illegal fishing and smuggling based on abnormal behavior \cite{fiorini2016}. While AIS data is often collected by land-based stations, a system based on buoys has been proposed to improve the offshore collection range of AIS data of up to 60\% \cite{lessing2006}.

A recent paper encourages the scientific community to apply machine learning to tackle climate change, stating that ``machine learning (ML) has been recognized as a broadly powerful tool for technological progress'' \cite{rolnick2019}. Clustering, i.e. identifying and grouping similar instances, is a well-known method within ML \cite{geron2019}.

Recently, clustering algorithms have been applied to AIS data for identifying abnormal behavior of ships \cite{yan2016}, determining vessel spatio-temporal co-occurrence patterns for improved maritime situational awareness \cite{wang2017}, combined with big data frameworks for increased performance of very large data sets \cite{chen2017}, route planning on the Yangtze River \cite{cao2018}, and classification of ship types in a port area \cite{zhou2019}.

This paper introduces a novel approach to determining offshore installation times by applying ML to AIS data from jackup vessels. We derive detailed time breakdowns of individual turbine installations, the variation within parks as well as the time spent in transit and docked in harbor. The new method requires no prior knowledge of individual turbine locations. Installation times and turbine locations can be inferred directly by applying a clustering algorithm to publicly available AIS data. These results can reduce the uncertainty of costs when planning future offshore projects and thereby reduce overall project costs \cite{Poulsen2017,Crown2012}.

Analyzing and understanding AIS data has for some years been a part of a larger tracking project carried out internally in Siemens Gamesa. Until now the results generated from AIS data has relied on traditional scripting, which required fundamental knowledge of the individual offshore sites. All in all the traditional scripting approach yields the same level of accuracy on results, but requires significantly more overhead in setting up and maintenance. The fact that the ML model is simpler to set up and maintain than traditional scripting was an eye opening experience contradicting the intuition of many project participants and therefore and experience worth sharing.

\newpage
\section{Methods}

\subsection{Data}
\begin{figure}[htp]
\centering
\includegraphics[width=\textwidth]{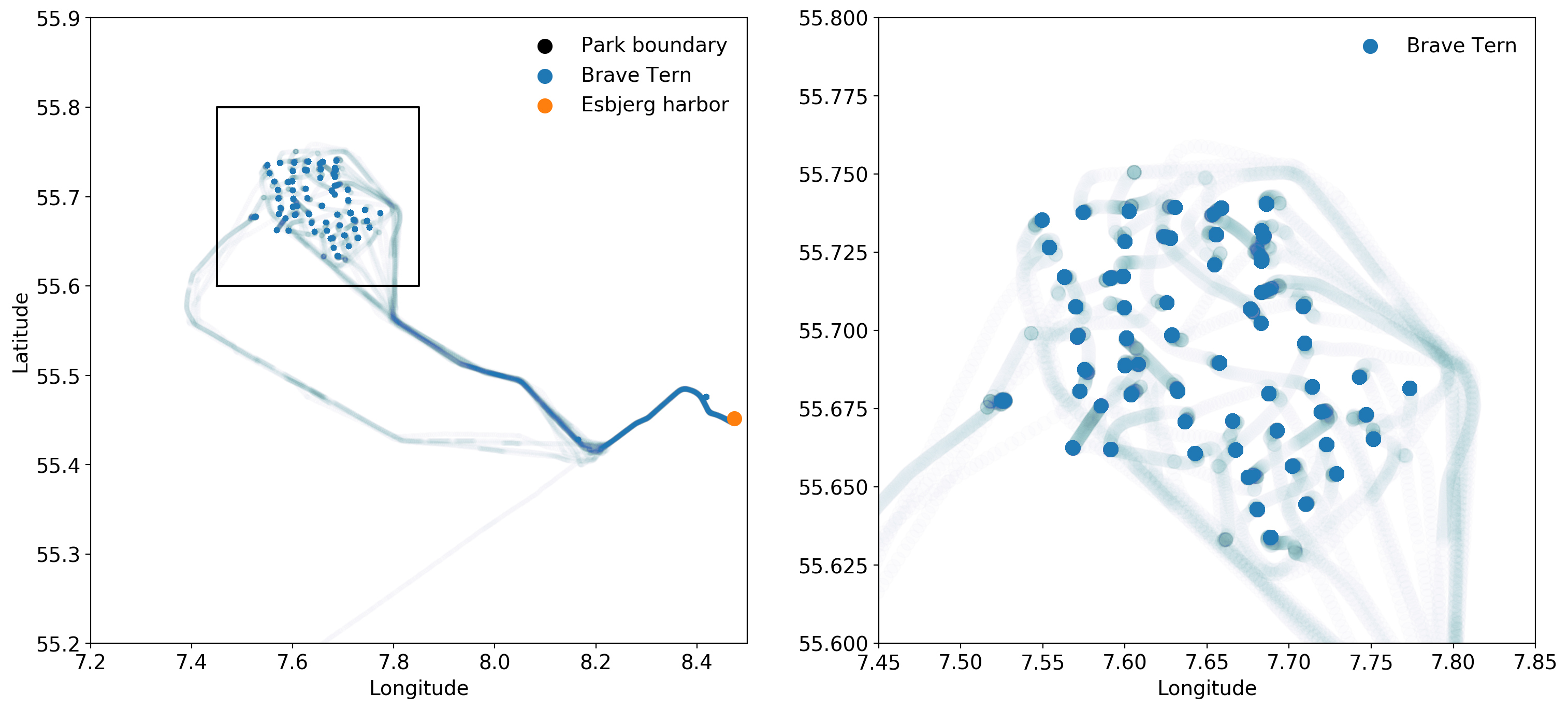}
\caption{Position data for the Brave Tern jackup vessel during installation of Horns Rev 3 from July 2018 to January 2019 (left). Zoomed in version corresponding to the black frame in the left part (right).}
\label{fig:ais}
\end{figure}

We exclusively use AIS data, which reports unique identification, position, course and speed for marine vessels. This data can either be collected directly from vessel transponder broadcasts or obtained from aggregators such as maritime authorities or data brokers. For Danish waters, historical AIS data is currently made publicly available for free by the Danish Maritime Authority \cite{ais}. For wind farms outside Danish waters, we obtain AIS data from MarineTraffic \cite{marinetraffic}. In this study, we only consider GPS position data (latitude, longitude). As an example, the left part of \fref{fig:ais} shows latitude and longitude extracted from AIS data for the Brave Tern jackup vessel during installation of Horns Rev 3 from July 2018 to January 2019. The right part of the figure shows the same data when zoomed in on the farm. The sampling frequency of the AIS data, publicly available from \cite{ais}, is approximately 0.1 Hz allowing for a very detailed tracking of the vessel.

An overview of the wind farms included in this study is shown in \tref{tab:farms} including number of turbines, geographical location, year when installation started and jackup vessel.

\begin{table}[htp]
\caption{The wind farms included in this study, their number of turbines, geographical location, year when installation started, and jackup vessel.}
\label{tab:farms}
\small
\centering
\begin{adjustbox}{center}
\begin{tabular}{lrlll}
\toprule
{} &  Turbines & Country &  Year &          Vessel \\
\midrule
Horns Rev 3                &        49 &      DK &  2018 &      Brave Tern \\
Arkona                     &        60 &      DK &  2018 &  Sea Challenger \\
Butendiek                  &        80 &      DE &  2014 &       Bold Tern \\
Dudgeon                    &        67 &      UK &  2016 &  Sea Challenger \\
Gode Wind                  &        97 &      DE &  2015 &  Sea Challenger \\
Beatrice                   &        84 &      UK &  2018 &    Pacific Orca \\
Burbo Bank Extension       &        32 &      UK &  2016 &  Sea Challenger \\
Galloper (BT)              &        17 &      UK &  2017 &       Bold Tern \\
Galloper (PO)              &        39 &      UK &  2017 &    Pacific Orca \\
Race Bank                  &        91 &      UK &  2017 &   Sea Installer \\
Rentel                     &        42 &      BE &  2018 &   Sea Installer \\
Westermost Rough           &        35 &      UK &  2014 &  Sea Challenger \\
Walney Extension (Siemens) &        47 &      UK &  2017 &          Scylla \\
Walney Extension (Vestas)  &        40 &      UK &  2017 &          Scylla \\
Hohe See (Brave Tern)      &        39 &      DE &  2019 &      Brave Tern \\
Hohe See (Blue Tern)       &        32 &      DE &  2019 &       Blue Tern \\
\bottomrule
\end{tabular}

\end{adjustbox}
\end{table}

\newpage
\subsection{Clustering}
The current state-of-the-art method for determining offshore wind installation times entails dividing the time interval from start to finish of the entire farm by the number of turbines \cite{Lacalarantegui2018}. This method provides an average installation time, which includes the time spent in transit between the wind farm and harbor as well as the time spent in the harbor. The use of AIS data to determine installation times of offshore wind turbines was proposed in \cite{Lacalarantegui2018}.

The new method presented here uses a ML method to cluster the GPS coordinates extracted from AIS data of jackup vessels. We are able to automatically identify installation times for individual turbines, which provides both an overall average, but also the distribution of installation times. In addition, we can identify the time spent in transit and the time spent in harbor, thus enabling a much more detailed description of the entire installation process.

We seek to determine installation times exclusively from AIS data since this is often the only data that is readily available to the public. Coordinates of individual turbines are generally not available. This can be remedied by applying a clustering algorithm to the GPS location data provided in the AIS broadcasts from every jackup vessel. Clustering of this data allows us to determine individual turbine coordinates and subsequently individual installation times.

Determining turbine locations is done using the K-means clustering algorithm as implemented in scikit-learn \cite{scikit-learn}. Briefly, it divides a set of observations into $k$ clusters by minimizing
\begin{equation}\label{eq:clustering}
\min_{\vec{S}} \sum_{i=1}^k\sum_{\vec{x}\in S_i} || \vec{x} - \bm{\mu}_i||^2,
\end{equation}
where $S_i$ is the subset of the data assigned to cluster $i$ and $\bm{\mu}_i$ the mean of this subset. Generally, the vector $\vec{x}$ can be n-dimensional. In this case it has just two dimensions: longitude and latitude. When applying the clustering algorithm to each of the wind farms in \tref{tab:farms}, the number of clusters $k$ is set to the number of actual turbines within the farm plus a few extra to account for the paths to and from the farm. The purpose of the extra clusters is to capture unnecessary data points so they do not impact the desired clusters. The extra clusters have been tweaked manually for each farm based on visual inspection and are automatically discarded during the subsequent process of determining installation times for each turbine. For a detailed description of clustering methods and the K-means algorithm, see \cite{geron2019}.

\begin{figure}[ht]
\centering
\includegraphics[width=.6\textwidth]{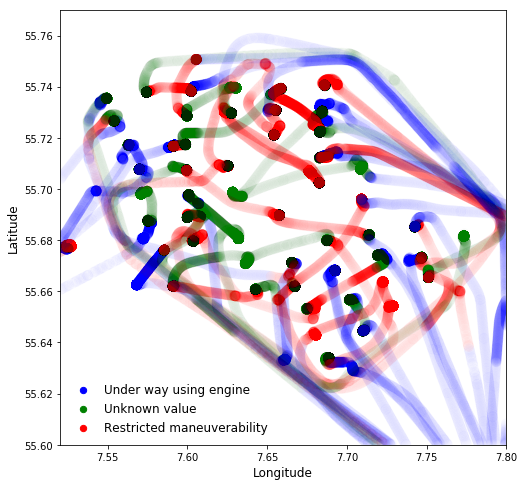}
\caption{Navigational status according to AIS for Brave Tern during installation of Horns Rev 3.}
\label{fig:nav-status}
\end{figure}

The AIS data includes a signal called \emph{Navigational Status}. This signal reports whether the vessel is moving, anchored etc. This signal could potentially be used to determine the location of turbines. However, the crew manually reports this signal, and therefore it is prone to error. \fref{fig:nav-status} shows the GPS coordinates for Brave Tern during installation of Horns Rev 3 as in \fref{fig:ais}. In this figure each point has been colored by its Navigational Status. The figure clearly shows that the Navigational Status cannot be relied upon for detecting when the vessel is installing a turbine.

\subsection{Identifying installation times}
Having identified turbine locations by clustering the AIS data we determine the installation time of each turbine based on the AIS data assigned to each cluster. We discard all positions that are further than 100 meters from the cluster center. The installation time is then determined as the time starting from the vessel entering this 100 meter radius of the turbine until it leaves. We observe a few cases of what appears to be more than one installation per turbine. These individual time segments are summed to one cumulative installation time per turbine location.

To account for low time resolution and missing data, we determine the uncertainty of the identified installation time by calculating the time interval from the first data point just before the vessel enters the 100 meter radius until the first data point just after the vessel has left the 100 meter radius. This 100 meter radius has been chosen with the criteria to be as small as possible while taking the typical size of a jackup vessel into account.

\newpage
\section{Results}
We apply the new method based on ML to 13 offshore wind farms in Danish, German and British waters as listed in \tref{tab:farms}. In the following subsections we first identify individual turbine locations. From these we determine individual installation times, time in transit and time in harbor. Finally, using wind speed data we determine the effect of weather conditions on installation times.

\subsection{Clustering}
\begin{figure}[ht]
\centering
\includegraphics[width=\textwidth]{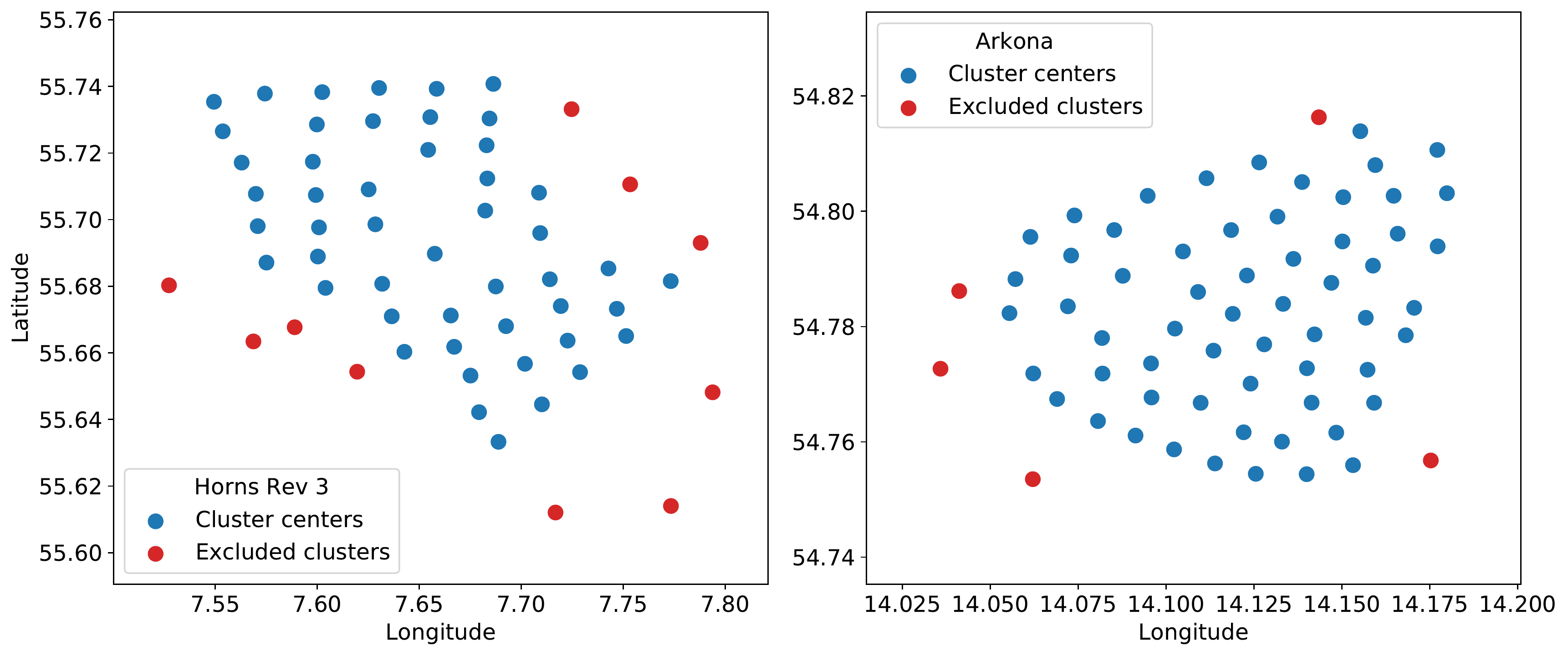}
\caption{Clustering of AIS data for Brave Tern installing Horns Rev 3 (left) and Sea Challenger installing Arkona (right).}
\label{fig:clustering}
\end{figure}

The results of applying the K-means clustering algorithm defined in \eqref{eq:clustering} to two wind farm installations in Danish waters are shown in \fref{fig:clustering}. The results for Brave Tern installing Horns Rev 3 and Sea Challenger installing Arkona are shown in the left and right panel, respectively. The figure shows the identified turbine locations from clustering AIS data in blue and additional clusters capturing the vessel's path to and from the wind farm in orange, which the algorithm has automatically marked for exclusion. The exclusion is based on the amount of data points within a cluster. The extra clusters contain much fewer data points for when the vessel is moving to and from the farm compared to being stationary during an installation.

\subsection{Installation times}
\begin{figure}[t]
\centering
\includegraphics[width=\textwidth]{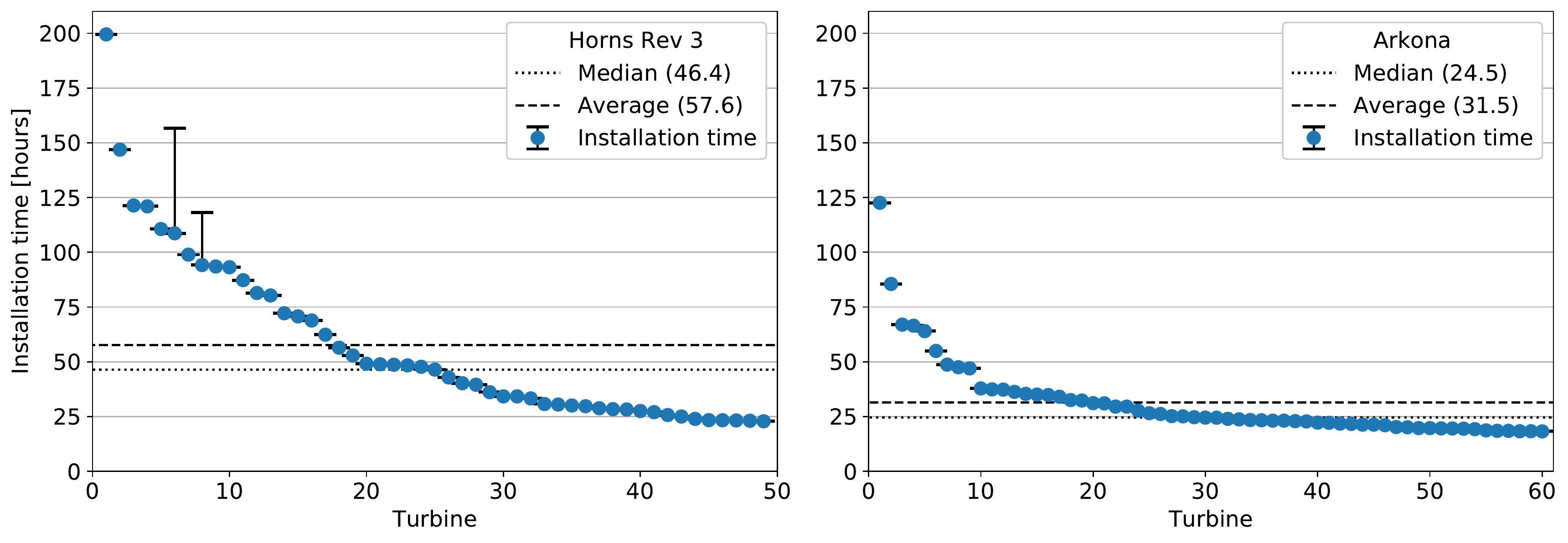}
\caption{Installation times sorted by duration for Horns Rev 3 (left) and Arkona (right).}
\label{fig:installations}
\end{figure}

The installation times resulting from the clusters shown in \fref{fig:clustering} are shown in \fref{fig:installations}. The left panel represents Horns Rev 3 and the right panel Arkona, where the identified installations are sorted by duration.

The errorbars indicate the uncertainty in identifying the duration of each installation. This uncertainty depends on the temporal resolution and completeness of available AIS data. The two installations with a high uncertainty in Horns Rev 3 are caused by gaps in the AIS time series. For the remaining examples shown here the uncertainties are almost non-existent. This is due to the fact that the AIS data collected from \cite{ais} generally has a high sampling frequency.

\tref{tab:stats} shows statistics on the identified installation times for each farm. We report the average, standard deviation, minimum, median and maximum times in hours. The coverage percentage is calculated as the number of identified turbine installations as a fraction of the actual number of turbines. These results are much more detailed than the current state-of-the-art of only reporting averages \cite{Lacalarantegui2018}. For all farms the median installation time is lower than the average, which is caused by a few outliers with a very long installation time. For most farms we are able to identify all turbines. The cases of missing turbines are due to missing data. The very high maximum installation time of 588 hours for Brave Tern at Hohe See is caused by 3 segments at the same location. They are each 295, 199 and 94 hours.

\begin{table}[t]
\caption{Statistics on the identified installation segments for each farm. We report the average, standard deviation, minimum, median and maximum times in hours. The coverage percentage is calculated as the number of identified turbine locations as a fraction of the actual number of turbines.}
\label{tab:stats}
\small
\centering
\begin{adjustbox}{center}
\begin{tabular}{lrrrrrrr}
\toprule
{} &  avg. &  s.d. &   min & median &    max & turbines & coverage [\%] \\
\midrule
Horns Rev 3                &  57.6 &  38.1 &  22.8 &   46.4 &  199.5 &       49 &           100 \\
Arkona                     &  31.5 &  18.4 &  18.2 &   24.5 &  122.6 &       60 &           100 \\
Butendiek                  &  53.7 &  53.8 &  16.7 &   25.3 &  303.4 &       80 &            96 \\
Dudgeon                    &  57.0 &  33.4 &  25.9 &   43.2 &  175.4 &       67 &            99 \\
Gode Wind                  &  39.5 &  44.4 &  15.4 &   22.2 &  242.0 &       97 &            94 \\
Beatrice                   &  66.1 &  62.7 &  20.1 &   38.3 &  452.9 &       84 &            94 \\
Burbo Bank Extension       &  50.4 &  33.2 &  23.9 &   36.8 &  176.5 &       32 &           100 \\
Galloper (BT)              &  93.2 &  74.6 &  29.6 &   83.4 &  319.9 &       17 &           100 \\
Galloper (PO)              &  84.8 &  43.3 &  29.0 &   76.8 &  214.6 &       39 &           100 \\
Race Bank                  &  37.2 &  16.7 &  19.3 &   31.6 &   87.2 &       91 &            89 \\
Rentel                     &  39.0 &  15.5 &  21.1 &   34.2 &   83.5 &       42 &           100 \\
Westermost Rough           &  52.4 &  30.7 &  22.9 &   41.1 &  126.6 &       35 &            83 \\
Walney Extension (Siemens) &  46.0 &  42.5 &  19.2 &   26.2 &  228.6 &       47 &           100 \\
Walney Extension (Vestas)  &  47.2 &  33.3 &  20.0 &   35.7 &  144.1 &       40 &           100 \\
Hohe See (Brave Tern)      &  54.3 &  90.7 &  20.8 &   32.7 &  588.1 &       39 &           100 \\
Hohe See (Blue Tern)       &  53.5 &  34.4 &  23.7 &   36.7 &  144.2 &       32 &           100 \\
\bottomrule
\end{tabular}

\end{adjustbox}
\end{table}

\begin{figure}[t]
\centering
\includegraphics[width=\textwidth]{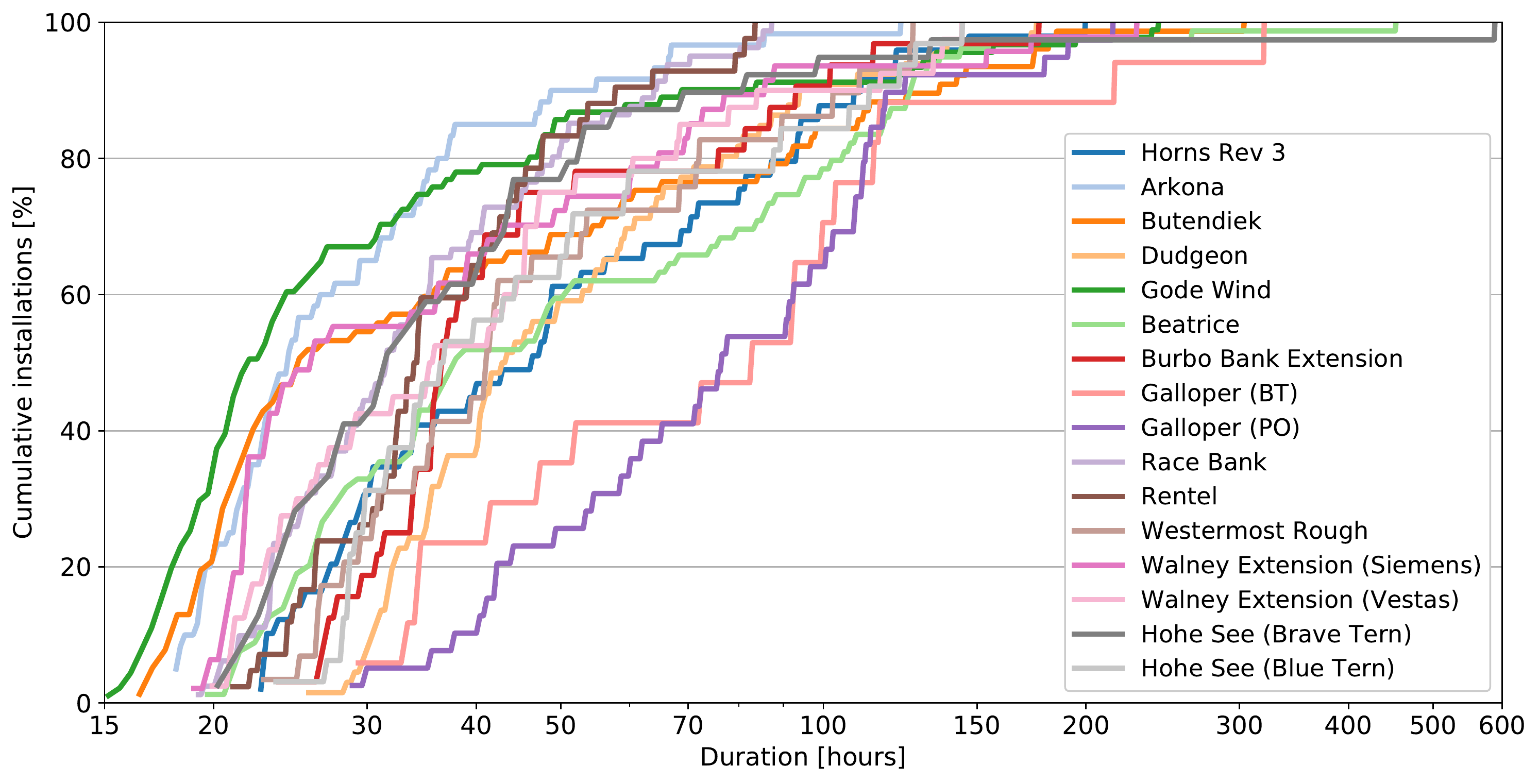}
\caption{Cumulative histogram of individual installation times compared between selected wind farms.}
\label{fig:cumulative}
\end{figure}

\fref{fig:cumulative} shows a comparison of the distribution of installation times across all wind farms. We see that most installation times are well below 100 hours and even below 50 hours, while there is a very small number of extreme cases with installation times of several hundred hours as shown also in \tref{tab:stats}. The figure shows a cumulative histogram of turbine installations per wind farm. As an example, it shows that for Gode Wind about 40\% of the installations each took less than 20 hours and about 80\% of the installations each took less than 40 hours. The steeper the curve, the lower the variation in installation times. The two leftmost curves for Gode Wind and Arkona are good examples of short installation times. On the other hand the two curves to the right for Bold Tern and Pacific Orca installing Galloper show a large variations in installation times. Note that the x-axis is logarithmic and that all installations have been identified automatically, so they have not been manually validated individually.

Similar to \fref{fig:cumulative}, it is possible to compare the distribution of installation times between jackup vessels. However, individual installation times will vary based on factors such as the size and weight of the installed components, weather conditions, seabed, and the fact that the different jackup vessels have different lifting capacities. Additionally, the total installation time of the farm depends on the number of turbines the jackup vessel can carry per trip. Due to these factors such benchmarking should be done with caution.

In addition to identifying installation times, we also identify the time spent in harbor by the jackup vessel. We use the same method as when identifying turbine locations and installation times. In the case of Horns Rev 3, Brave Tern docked in Esbjerg harbor 21 times with an average docking time of 82.5 hours for a total of 1732 hours spent in harbor. These numbers are reported in \tref{tab:harbor}. The docking time should reflect the number of turbines carried per trip. There is one occurrence of Brave Tern spending almost two full weeks docked. Such long docking times are not uncommon and can be caused by a combination of bad weather conditions, vessel repairs/upgrades, and shortage of turbine components from production.

For Horns Rev 3 we have considered a time interval from 00:00, July 1, 2018 to 12:30, January 21, 2019. This interval corresponds to 4932.5 hours. Of these hours 2821 was spent installing turbines, 1732 was spent in harbor, and 379.5 were spent in transit between Horns Rev 3 and Esbjerg Harbor. See \tref{tab:times} for a comparison of the time spent and the respective percentages. The time spent in transit is calculated as the total time minus the installation and harbor times. It depends on the distance of the wind farm from shore and available shipping lanes between the farm and harbor. Distinguishing between these different classes of time segments enables detailed modeling of offshore wind installation costs \cite{costModeling}.

Following the approach in \cite{Lacalarantegui2018}, the average installation time per turbine for Horns Rev 3 is calculated by dividing the total time by the number of turbines, which results in 100.6 hours per turbine. This is an increase of 87.9\% compared with the average time reported in \tref{tab:stats}, because the approach in \cite{Lacalarantegui2018} does not take into account the time spent in transit and harbor. However, it is important to estimate both how much time is spent installing turbines and how much time is spent docked. This is due to the possibility of the rate of a jackup vessel being variable depending on the amount of crew required, which is reduced when docked for longer periods.

\begin{table}[t]
\caption{Statistics on the duration in hours of the 21 identified harbor segments for Horns Rev 3.}
\label{tab:harbor}
\small
\centering
\begin{adjustbox}{center}
\begin{tabular}{rrrrrr}
\toprule
segments &  avg. &  s.d. &   min & median &    max \\
\midrule
      21 &  82.5 &  72.3 &  15.5 &   62.3 &  317.4 \\
\bottomrule
\end{tabular}

\end{adjustbox}
\end{table}

\begin{table}[t]
\caption{Overview of the amount of time spent in harbor, transit and installation for Brave Tern during installation of Horns Rev 3.}
\label{tab:times}
\small
\centering
\begin{tabular}{lrrrrr}
\toprule
 &  transit & installation & harbor & total\\
\midrule
Time [hours] & 379.5 & 2821 & 1732 & 4932.5  \\
Share [\%] & 7.7 & 57.2 & 35.1 & 100 \\
\bottomrule
\end{tabular}
\end{table}

\begin{figure}[ht]
\centering
\includegraphics[width=\textwidth]{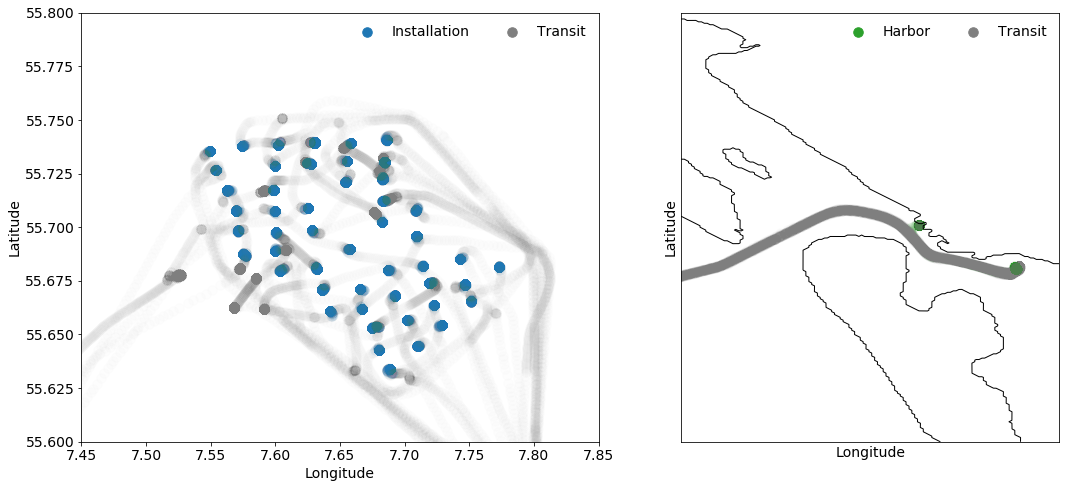}
\caption{Position data for Brave Tern during installation of Horns Rev 3 from July 2018 to January 2019. The data has been colored as the result of identifying installation, harbor and transit segments. Inferred turbine locations are shown in the left part and the right part shows docking locations at Esbjerg Harbor.}
\label{fig:ais-segment}
\end{figure}

Similar to \fref{fig:ais}, \fref{fig:ais-segment} shows AIS data for Horns Rev 3. In this figure the data has been colored as the result of identifying all installation, harbor and transit segments during the installation of Horns Rev 3. The left part shows the turbine locations inferred by applying the clustering algorithm to the AIS data. The right part shows the two docking locations in Esbjerg Harbor where loading of Brave Tern took place.

\subsection{Effect of wind speed on installation time}
\begin{figure}[ht]
\centering
\includegraphics[width=.8\textwidth]{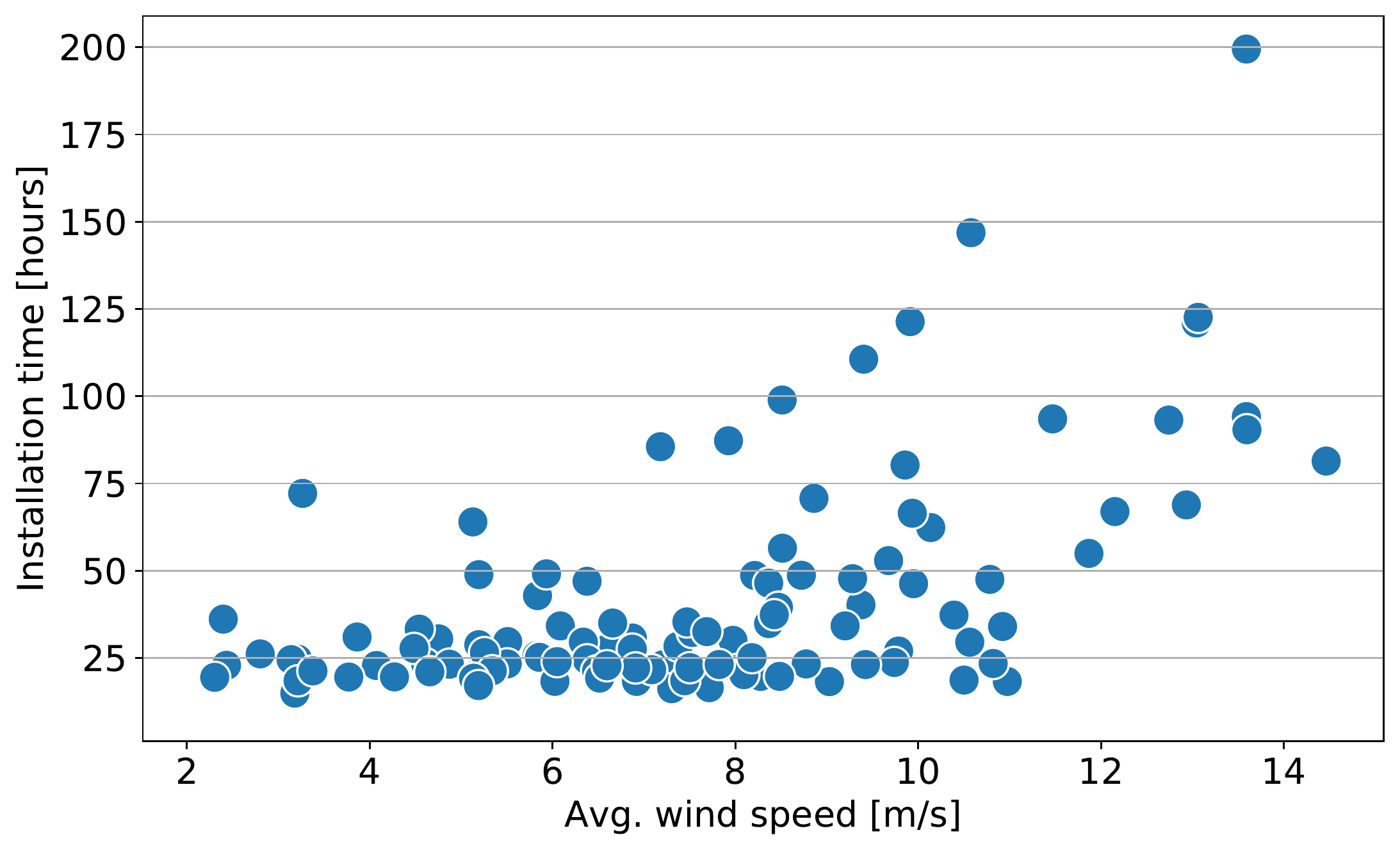}
\caption{Relation between wind speed and installation time.}
\label{fig:wind}
\end{figure}

Weather conditions such as wind speed and wave height have an effect on installation time \cite{smart2016,Vis2016}. We briefly explore this in \fref{fig:wind}, which shows the installation time as a function of average wind speed during installation of Horns Rev 3 and Arkona.

This figure shows that there is a relation between average wind speed and installation time and that it is not simply linear. On average, the installation time tends to increase with increasing average wind speed. More importantly, the variation in installation time increases heavily with increasing wind speed. This shows that the relation between wind speed and installation time is nonlinear. However, further analysis is necessary in order to generalize between wind farm projects.

\section{Conclusion}
Based on the identified time intervals for installation, transit and harbor we determined the performance of the installation process for each wind farm. These results allow planners of future wind farm projects to base their cost estimates of installation times on generalizations of the results presented here. Increasing accuracy of cost estimates leads to reduced financial risk and thereby lowering the overall project costs. This analysis also enables identifying particularly efficient installations times in order to learn from them. Additionally, these results allow us to compare installation times between projects and benchmark the performance of different jackup vessels. However, such comparisons should be done with caution since the different wind farms use turbines of varying sizes and weight, which influence the expected installation times. Further, the different jackup vessels have varying lifting capacities limiting the availability of vessels for turbines of increasing size. Bad weather conditions also have an effect: high wind speeds or wave heights might cause delays in jacking up and lifting. Additionally the quality of the seabed, in combination with the weight of the turbine components, significantly impacts the jackup and jack-down times.

Due to the number of factors affecting installation times it might not be straightforward to generalize the results on installation times in this study to future wind farms. However, the method presented here allows turbine manufacturers to continuously monitor their installation processes. This enables a continuous feedback loop where previous estimates of installation times can be evaluated. Such comparisons can in turn be used to improve the estimates for future offshore wind projects in an iterative way. This will lead to higher accuracy in predictions and, thus, lower project costs due to increasingly reduced risk.

\section*{Acknowledgments}
We thank Jacob Bjerre for helpful discussions and for providing access to wind speed data from internal Siemens Gamesa databases.

\bibliographystyle{unsrtDOI}
\bibliography{references}

\begin{thebibliography}{10}
\expandafter\ifx\csname href\endcsname\relax
  \def\href#1#2{#2} \def\path#1{#1}\fi

\bibitem{WindEUStats}
WindEurope.
\newblock Wind in power 2017.
\newblock Technical report, WindEurope, 2018.
\newblock Available at
  \url{https://windeurope.org/about-wind/statistics/european/wind-in-power-2017/}.

\bibitem{WindEU2020}
WindEurope.
\newblock {Wind energy in Europe: Outlook to 2020}.
\newblock Technical report, {WindEurope}, 2017.
\newblock Available at
  \url{https://windeurope.org/about-wind/reports/wind-energy-in-europe-outlook-to-2020/}.

\bibitem{irena2019}
International Renewable Energy~Agency (IRENA).
\newblock {Renewable Power Generation Costs in 2018}, 2019.
\newblock Available at
  \url{https://www.irena.org/publications/2019/May/Renewable-power-generation-costs-in-2018}.

\bibitem{vattenfall2018}
Vattenfall.
\newblock {Vattenfall wins tender Hollandse Kust Zuid}, 2018.
\newblock Available at
  \url{https://group.vattenfall.com/press-and-media/news--press-releases/pressreleases/2018/vattenfall-wins-tender-hollandse-kust-zuid}.

\bibitem{orsted2018}
Ørsted.
\newblock {Ørsted wins 551.75MW in German offshore wind auction}, 2018.
\newblock Available at
  \url{https://orsted.com/en/Company-Announcement-List/2018/04/1731313}.

\bibitem{Ahn2017}
Dang Ahn, Sung chul Shin, Soo young Kim, Hicham Kharoufi, and Hyun cheol Kim.
\newblock Comparative evaluation of different offshore wind turbine
  installation vessels for korean west–south wind farm.
\newblock {\em International Journal of Naval Architecture and Ocean
  Engineering}, 9(1):45 -- 54, 2017, \href
  {https://doi.org/10.1016/j.ijnaoe.2016.07.004}
  {\path{doi:10.1016/j.ijnaoe.2016.07.004}}.

\bibitem{Mou2010}
Jun~Min Mou, Cees van~der Tak, and Han Ligteringen.
\newblock Study on collision avoidance in busy waterways by using ais data.
\newblock {\em Ocean Engineering}, 37(5):483 -- 490, 2010, \href
  {https://doi.org/10.1016/j.oceaneng.2010.01.012}
  {\path{doi:10.1016/j.oceaneng.2010.01.012}}.

\bibitem{ais}
Danish~Maritime Authority.
\newblock {AIS} data.
\newblock
  \url{https://www.dma.dk/SikkerhedTilSoes/Sejladsinformation/AIS/Sider/default.aspx},
  Accessed April 2019.

\bibitem{marinetraffic}
MarineTraffic.
\newblock \url{https://www.marinetraffic.com}, Accessed April 2019.

\bibitem{fiorini2016}
Michele Fiorini, Andrea Capata, and Domenico~D. Bloisi.
\newblock {{AIS Data Visualization for Maritime Spatial Planning (MSP)}}.
\newblock {\em International Journal of e-Navigation and Maritime Economy},
  5:45 -- 60, 2016, \href {https://doi.org/10.1016/j.enavi.2016.12.004}
  {\path{doi:10.1016/j.enavi.2016.12.004}}.

\bibitem{lessing2006}
P.~A. {Lessing}, L.~J. {Bernard}, B.~J. {Tetreault}, and J.~N. {Chaffin}.
\newblock {Use of the Automatic Identification System (AIS) on Autonomous
  Weather Buoys for Maritime Domain Awareness Applications}.
\newblock In {\em OCEANS 2006}, pages 1--6, Sep. 2006, \href
  {https://doi.org/10.1109/OCEANS.2006.307023}
  {\path{doi:10.1109/OCEANS.2006.307023}}.

\bibitem{rolnick2019}
David Rolnick, Priya~L. Donti, Lynn~H. Kaack, Kelly Kochanski, Alexandre
  Lacoste, Kris Sankaran, Andrew~Slavin Ross, Nikola Milojevic-Dupont, Natasha
  Jaques, Anna Waldman-Brown, Alexandra Luccioni, Tegan Maharaj, Evan~D.
  Sherwin, S.~Karthik Mukkavilli, Konrad~P. Kording, Carla Gomes, Andrew~Y. Ng,
  Demis Hassabis, John~C. Platt, Felix Creutzig, Jennifer Chayes, and Yoshua
  Bengio.
\newblock Tackling climate change with machine learning.
\newblock \href{https://arxiv.org/abs/1906.05433}{arXiv:1906.05433}, 2019.

\bibitem{geron2019}
Aurélien Géron.
\newblock {\em Hands-on Machine Learning with Scikit-Learn, Keras, and
  TensorFlow}.
\newblock O'Reilly Media, Inc., 2nd edition, 2019.

\bibitem{yan2016}
{Yan Li}, Y.~{Zhang}, and F.~{Zhu}.
\newblock {The method of detecting AIS isolated information based on clustering
  and distance}.
\newblock In {\em 2016 2nd IEEE International Conference on Computer and
  Communications (ICCC)}, pages 870--873, Oct 2016, \href
  {https://doi.org/10.1109/CompComm.2016.7924827}
  {\path{doi:10.1109/CompComm.2016.7924827}}.

\bibitem{wang2017}
Jiang Wang, Cheng Zhu, Yun Zhou, and Weiming Zhang.
\newblock {Vessel Spatio-temporal Knowledge Discovery with AIS Trajectories
  Using Co-clustering}.
\newblock {\em Journal of Navigation}, 70(6):1383–1400, 2017, \href
  {https://doi.org/10.1017/S0373463317000406}
  {\path{doi:10.1017/S0373463317000406}}.

\bibitem{chen2017}
Z.~{Chen}, J.~{Guo}, and Q.~{Liu}.
\newblock {DBSCAN Algorithm Clustering for Massive AIS Data Based on the Hadoop
  Platform}.
\newblock In {\em 2017 International Conference on Industrial Informatics -
  Computing Technology, Intelligent Technology, Industrial Information
  Integration (ICIICII)}, pages 25--28, Dec 2017, \href
  {https://doi.org/10.1109/ICIICII.2017.72}
  {\path{doi:10.1109/ICIICII.2017.72}}.

\bibitem{cao2018}
J.~{Cao}, M.~{Liang}, Y.~{Li}, J.~{Chen}, H.~{Li}, R.~W. {Liu}, and J.~{Liu}.
\newblock {PCA}-based hierarchical clustering of ais trajectories with
  automatic extraction of clusters.
\newblock In {\em 2018 IEEE 3rd International Conference on Big Data Analysis
  (ICBDA)}, pages 448--452, March 2018, \href
  {https://doi.org/10.1109/ICBDA.2018.8367725}
  {\path{doi:10.1109/ICBDA.2018.8367725}}.

\bibitem{zhou2019}
Yang Zhou, Winnie Daamen, Tiedo Vellinga, and Serge~P. Hoogendoorn.
\newblock {Ship classification based on ship behavior clustering from AIS
  data}.
\newblock {\em Ocean Engineering}, 175:176 -- 187, 2019, \href
  {https://doi.org/10.1016/j.oceaneng.2019.02.005}
  {\path{doi:10.1016/j.oceaneng.2019.02.005}}.

\bibitem{Poulsen2017}
Thomas Poulsen, Charlotte~Bay Hasager, and Christian~Munk Jensen.
\newblock The role of logistics in practical levelized cost of energy reduction
  implementation and government sponsored cost reduction studies: Day and night
  in offshore wind operations and maintenance logistics.
\newblock {\em Energies}, 10(4), 2017, \href
  {https://doi.org/10.3390/en10040464} {\path{doi:10.3390/en10040464}}.

\bibitem{Crown2012}
The~Crown Estate.
\newblock Offshore wind cost reduction pathways study, 2012.

\bibitem{Lacalarantegui2018}
Roberto Lacal-Arántegui, José~M. Yusta, and José~Antonio Domínguez-Navarro.
\newblock Offshore wind installation: Analysing the evidence behind
  improvements in installation time.
\newblock {\em Renewable and Sustainable Energy Reviews}, 92:133 -- 145, 2018,
  \href {https://doi.org/10.1016/j.rser.2018.04.044}
  {\path{doi:10.1016/j.rser.2018.04.044}}.

\bibitem{scikit-learn}
F.~Pedregosa, G.~Varoquaux, A.~Gramfort, V.~Michel, B.~Thirion, O.~Grisel,
  M.~Blondel, P.~Prettenhofer, R.~Weiss, V.~Dubourg, J.~Vanderplas, A.~Passos,
  D.~Cournapeau, M.~Brucher, M.~Perrot, and E.~Duchesnay.
\newblock Scikit-learn: Machine learning in {P}ython.
\newblock {\em Journal of Machine Learning Research}, 12:2825--2830, 2011.

\bibitem{costModeling}
Mark~J. Kaiser and Brian~F. Snyder.
\newblock {\em Offshore Wind Energy Cost Modeling: Installation and
  Decommissioning}.
\newblock Springer, 2012, \href {https://doi.org/10.1007/978-1-4471-2488-7}
  {\path{doi:10.1007/978-1-4471-2488-7}}.

\bibitem{smart2016}
Gavin Smart, Aaron Smith, Ethan Warner, Iver~Bakken Sperstad, Bob Prinsen, and
  Roberto Lacal-Arantegui.
\newblock {IEA Wind Task 26: Offshore Wind Farm Baseline Documentation}.
\newblock \href {https://doi.org/10.2172/1259255} {\path{doi:10.2172/1259255}}.

\bibitem{Vis2016}
Iris~F.A. Vis and Evrim Ursavas.
\newblock Assessment approaches to logistics for offshore wind energy
  installation.
\newblock {\em Sustainable Energy Technologies and Assessments}, 14:80 -- 91,
  2016, \href {https://doi.org/10.1016/j.seta.2016.02.001}
  {\path{doi:10.1016/j.seta.2016.02.001}}.

\end{thebibliography}

\end{document}